\pdfoutput=1

\documentclass[11pt]{article}

\usepackage[]{emnlp2021}

\usepackage{times}
\usepackage{latexsym}

\usepackage[T1]{fontenc}

\usepackage[utf8]{inputenc}

\usepackage{microtype}

\usepackage{booktabs,subcaption,amsmath}
\usepackage{tikz}
\usepackage{tikz-qtree}
\usepackage{enumitem} 
\usepackage{fancyvrb}
\usepackage{mathtools}
\usepackage{amsmath}
\usepackage{multirow}
\usepackage{algorithm}
\usepackage[noend]{algpseudocode}
\usepackage{mathtools}
\usepackage{booktabs}
\usepackage{wrapfig}
\usepackage{inconsolata}

\title{Span Pointer Networks for Non-Autoregressive \\ Task-Oriented Semantic Parsing}

\author{Akshat Shrivastava \quad\quad Pierce Chuang \quad\quad Arun Babu \quad\quad Shrey Desai \\ 
 \textbf{Abhinav Arora } \quad\quad \textbf{Alexander Zotov} \quad\quad \textbf{Ahmed Aly} \\
        Facebook \\
        \tt{\{akshats, pichuang, arbabu, shreyd},\\ \tt{abhinavarora, azotov, ahhegazy\}@fb.com}}

\begin{document}
\maketitle
\begin{abstract}
An effective recipe for building seq2seq, non-autoregressive, task-oriented parsers to map utterances to semantic frames proceeds in three steps: encoding an utterance $x$, predicting a frame's length $|y|$, and decoding a $|y|$-sized frame with utterance and ontology tokens. Though empirically strong, these models are typically bottlenecked by length prediction, as even small inaccuracies change the syntactic and semantic characteristics of resulting frames. In our work, we propose span pointer networks, non-autoregressive parsers which shift the decoding task from text generation to span prediction; that is, when imputing utterance spans into frame slots, our model produces endpoints (e.g., \texttt{[i, j]}) as opposed to text (e.g., ``6pm''). 
This natural quantization of the output space also provides consistency in the length prediction task, allowing our length predictor to be responsible for frame syntax and the decoder for frame syntax, creating a coarse-to-fine model.
We evaluate our approach on several task-oriented semantic parsing datasets. Notably, we bridge the quality gap between non-autogressive and autoregressive parsers, achieving 87 EM on TOPv2 \cite{chen-2020-topv2}. Furthermore, due to our more consistent gold frames, we show strong improvements in model generalization in both cross-domain and cross-lingual transfer in low-resource settings. Finally, due to our diminished output vocabulary, we observe 70\% reduction in latency and 83\% in memory at beam size 5 compared to prior non-autoregressive parsers. 

\end{abstract}

\section{Introduction}

\begin{figure}[t]
\centering
\includegraphics[scale=0.225]{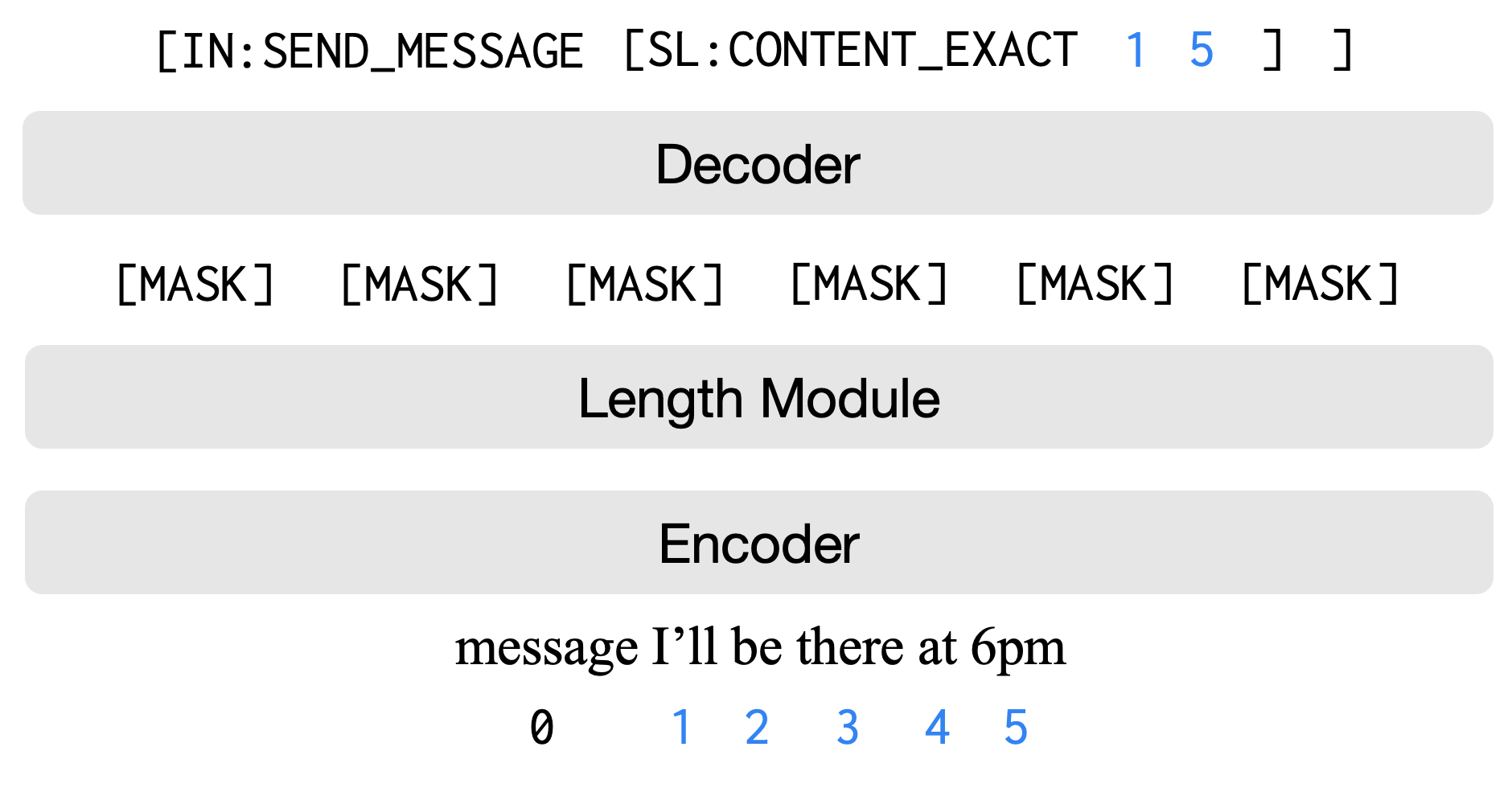}
\caption{ Illustration of a span pointer network for task-oriented semantic parsing based on seq2seq, non-autoregressive, mask-predict models \cite{ghazvininejad2019maskpredict}. In this example, we map the utterance $x$ ``message I'll be there at 6pm'' to the frame $y$ \texttt{[IN:SEND\_MESSAGE [SL:CONTENT\_EXACT 1 5 ] ]} where \texttt{[1, 5]} corresponds to the span ``I'll be there at 6pm''.  Our model operates in three stages: (1) an encoder consumes the utterance; (2) a length module creates $|y|\: \times$ \texttt{[MASK]} tokens corresponding to the frame's length; and (3) a decoder swaps each \texttt{[MASK]} token with an utterance or ontology (e.g., intent or slot) token. Unlike typical methodology, we model span endpoints, which empirically simplifies the parsing task.
}
\label{fig:span_pointer_model}
\end{figure}

Task-oriented conversational assistants typically first employ semantic parsers to map utterances to frames \cite{hemphill1990atis,coucke2018snips,gupta2018semantic,rongali2020don,decoupled}. Due to performance constraints in real-world deployments, recent work in task-oriented semantic parsing has shifted towards building seq2seq, non-autoregressive parsers optimized for both quality and latency \cite{Zhu2020DontPI,nar_semantic_parsing}. These models enforce strong independence assumptions during decoding, allowing frame components (e.g., ontology and utterance tokens) to be generated in parallel. However, the application of off-the-shelf, non-autoregressive algorithms to task-oriented semantic parsing is not trivial, often leading to brittle implementations with sub-optimal, opinionated components \cite{nar_semantic_parsing}.

One popular family of seq2seq, non-autoregressive, task-oriented parsers is based on the mask-predict algorithm \cite{ghazvininejad2019maskpredict}, which operates in three steps: encoding, length prediction, and decoding.
Unlike machine translation, task-oriented semantic parsing does not benefit much from iterative refinement; therefore, frames are typically generated in one step. However, this increases the burden of upstream components, namely placing a major bottleneck on the length prediction module. Therefore, even off-by-one length errors can change the syntactic and semantic characteristics of resulting frames.

In this work, we create \textbf{span pointer networks} which model a seq2seq, non-autoregressive parsing task centered around span prediction as opposed to text generation. Figure \ref{fig:span_pointer_model} illustrates an application of our model; when parsing the utterance ``message I'll be there at 6pm'', our model produces span endpoints \texttt{[1, 5]} as opposed to exact text ``I'll be there at 6pm'' in the appropriate leaf slot. As a result, our length module implicitly predicts the syntax of a frame (i.e., how many intents and slots does the utterance have?) and the decoder resolves leaf arguments with span endpoints (i.e., what utterance spans should be included in the frame?). This additionally creates a parallel among our approach and coarse-to-fine \cite{Dong2018CoarsetoFineDF} modeling as our length predictor predicts a general structure that our decoder further refines. Despite this shift in behavior, our approach is largely compatible with the typical mask-predict methodology, as we largely modify the frame representation and model architecture to be span-based.

We evaluate span pointer networks on three axes: quality, generalizability, and resources. First, we benchmark exact match (EM) on multiple task-oriented semantic parsing datasets, including TOPv2 \cite{chen-2020-topv2} and TOP \cite{gupta2018semantic}. Our non-autoregressive parser is competitive, achieving 87 EM on TOPv2, and matches or exceeds autoregressive parsers in many cases. 
Second, we evaluate generalizability by setting up cross-domain and cross-lingual transfer experiments. Our non-autoregressive outperforms strong baselines, notably outperforming autoregressive parsers by +15 EM when averaged across 5 zero-shot multilingual settings.
Third, due to the restricted decoder vocabulary, we measure resource savings, in particular, latency and memory usage. Compared to non-autoregressive models our parser achieves 70\% reduction in latency and 83\% reduction in memory with a beam size of 5. 

To summarize, our contributions are: (1) We propose span pointer networks, non-autoregressive parsers which use span prediction as opposed to text generation; (2) We evaluate several aspects of span pointer networks, conducting a series of experiments around quality, generalizability, and resources; (3) We empirically show improvements on all three axes, notably outperforming both non-autoregressive and autoregressive parsers.
\section{Background and Related Work}

Task-oriented semantic parsing broadly consists of mapping textual utterances to structured frames \cite{gupta2018semantic, decoupled, chen-2020-topv2, rongali2020don, li-2020-mtop}. Frames are structured semantic representations of utterances, and are comprised of both ontology tokens (e.g., intents and slots) and utterance tokens. For example, we can map the utterance ``message I'll be there at 6pm'' into a frame \texttt{[IN:SEND\_MESSAGE [SL:CONTENT\_EXACT \textrm{I'll be there at 6pm} ] ]}, which has several components: (1) a ``send message'' intent; (2) a ``content exact'' slot; and (3) a span ``I'll be there at 6pm'', which is an argument of the ``content exact'' slot. Here, intents and slots are akin to functions and arguments, respectively, in an API call.

Modern parsers are typically based on seq2seq transformers which encode utterances and decode frames. These parsers, by virtue of being seq2seq models, can be learned either in an autoregressive or non-autoregressive fashion. Autoregressive parsers generate frame components in sequence, while non-autoregressive parsers generate frame components in parallel. Both paradigms have their advantages and disadvantages, primarily trading-off latency and quality: while non-autoregressive models are optimized for inference and suitable in production settings, they are often qualitatively worse than autoregressive models due to strong independence assumptions during generation \cite{gu2017non, lee2018, ghazvininejad2019maskpredict, kasai2020parallel, nar_semantic_parsing}. 

Our work focuses on improving the performance of one family of non-autoregressive parsers, namely those based on the mask-predict algorithm \cite{ghazvininejad2019maskpredict}. These parsers operate in three stages: (1) on the source-side, an encoder consumes an utterance $x$; (2) using its representations, a length module predicts a frame's length $|y|$ and creates $|y|\: \times$ \texttt{[MASK]} tokens; and (3) a length-conditioned decoder produces a frame $y$ with ontology and utterance tokens. Frames are typically generated in one step as, unlike machine translation, task-oriented semantic parsing does not benefit much from iterative refinement \cite{nar_semantic_parsing}. Consequently, this increases the burden of accuracy on upstream components, especially placing a major bottleneck on the length prediction module. Even off-by-one length errors can dramatically change resulting frames, which is one of the drivers behind the quality gap between non-autoregressive and autoregressive models.
\section{Span Pointer Networks}

In this section, we introduce span pointer networks: seq2seq, non-autoregressive parsers which model a span-based, task-oriented semantic parsing task. Our approach is based on the mask-predict algorithm, which requires an encoder, length module, and decoder to map utterances to frames. However, our core idea is shifting from \textbf{text generation} to \textbf{span prediction} when creating leaf arguments in frames; that is, replacing text spans (e.g., ``I'll be there at 6pm'') with index spans (e.g., \texttt{[1, 5]}). Text generation and span prediction are \textit{functionally} identical; though span prediction requires converting index spans to text spans post-hoc, both resolve to the same frame. However, text generation and span prediction are \textit{technically} different, as they present distinct targets for non-autoregressive parsing. Note that our approach is different from \citet{pasupat-etal-2019-span}, who develop a CKY-based parser which maps spans to labels; in contrast, we explore span prediction with seq2seq modeling.

To build intuition for how these paradigms affect mask-predict parsers, in particular, consider the utterance ``message I'll be there at 6pm'' and frames \texttt{[IN:SEND\_MESSAGE [SL:CONTENT\_EXACT \textrm{I'll be there at 6pm} ] ]} and \texttt{[IN:SEND\_MESSAGE [SL:CONTENT\_EXACT 1 5 ] ]}, one with text spans (requiring text generation) and one with index spans (requiring span prediction). \textbf{For accurate parsing, a length module needs to understand both frame syntax \textit{and} semantics with text generation, but \textit{only} frame syntax with span prediction.} Specifically, with text generation, the length module outputs 9 \texttt{[MASK]} tokens, which requires it to implicitly guess the entire frame---both its syntactic structure and semantic arguments---before decoding. In contrast, with span prediction, the length module outputs 6 \texttt{[MASK]} tokens, which requires it to implicitly guess syntactic structure but leave semantic arguments to the decoder, which will subsequently resolve the exact span indices.

To make this argument more clear, consider the frames \texttt{[SL:ARTIST \textrm{Beyonce} ]} and \texttt{SL:LOCATION \textrm{Seattle, Washington} ]}: these frames both have the same length despite being semantically distinct, so during fine-tuning, the length module learns syntax and the decoder learns semantics. We view this type of span prediction as analogous to coarse-to-fine modeling \cite{Dong2018CoarsetoFineDF}; the length module predicts a coarse-grained frame structure and the decoder infills incomplete parts (e.g., leaf arguments) with fine-grained details.

Below, we elaborate on the core components required to implement span pointer networks: a span-based \textbf{frame representation} and the \textbf{model architecture}.

\subsection{Frame Representation}
\label{sec:index_span}

\begin{table*}
\small
\centering
\begin{tabular}{p{2.25cm}p{10cm}}
\toprule
Form & Utterance / Decoupled Frame \\
\midrule
 & message I'll be there at 6pm \\
Canonical & \texttt{[IN:SEND\_MESSAGE [SL:CONTENT [} I'll be there at 6pm \texttt{] ] ]} \\
Index & \texttt{[IN:SEND\_MESSAGE [SL:CONTENT [ 1 2 3 4 5 ] ] ]} \\
Span & \texttt{[IN:SEND\_MESSAGE [SL:CONTENT [ 1 5 ] ] ]} \\
\bottomrule
\end{tabular}
\caption{ \textbf{Comparison of the canonical, index, and span forms of the decoupled frame representation.} Given the utterance ``message I'll be there at 6pm'', we present several decoupled representation forms, which each convey the same information, albeit in a different manner. Our span-based, non-autoregressive parser uses the span form.}
\label{tab:decoupled-comparison}
\end{table*}

Our span-based frame representation is a variant of the \textbf{decoupled} frame representation \cite{decoupled}, following prior work in task-oriented semantic parsing \cite{decoupled, li-2020-mtop}. This representation requires frames to mimic tree structures comprised of ontology tokens (e.g., intents and slots) and utterance tokens, where, critically, utterance tokens appear only as leaf arguments to slots. However, seq2seq parsers fine-tuned on decoupled frames typically require text generation, as leaf arguments consist of text spans. We refer to the original decoupled representation as the \textbf{canonical form} and propose two successive modifications, the \textbf{index form} and \textbf{span form} which, together, enable span prediction. A side-by-side view of these forms is depicted in Table \ref{tab:decoupled-comparison}, and we elaborate on the non-canonical forms below:
\begin{enumerate}
    \item \textbf{Index Form:} We capitalize on the closed nature of the semantic parsing task, replacing utterance tokens with index tokens which instead ``point'' to utterance positions \cite{rongali2020don}. Here, the definition of an index is contingent on the tokenization algorithm used and requires careful preprocessing of utterances and frames. However, when used consistently, index form yields several advantages. Transformers can model index-based frames out-of-the-box by leveraging the positional embeddings of utterance tokens \cite{transformer}. Furthermore, we can significantly restrict the decoder vocabulary as the maximum sequence length (e.g., 100) is typically several orders of magnitude smaller than the size of popular subword vocabularies (e.g., 50K BPE \cite{sennrich2015neural, roberta} and 250K SentencePiece \cite{kudo-richardson-2018-sentencepiece, conneau-2020-xlmr}).
    \item \textbf{Span Form:} From the index form, where utterance tokens are represented as index tokens, we create the span form, where the frame representation is simplified by collapsing index tokens into spans. Put simply, we modify leaf arguments to be index endpoints corresponding to utterance spans. To represent these spans, we use index tokens instead of text tokens primarily to eliminate ambiguity: if an utterance contains multiple instances of the same text token, it might be unclear which instance a span is referring to, but assigning tokens unique indices resolves this issue. In comparison to index form, one advantage of span form is that it makes the output space denser, improving the performance of the length module; for example, both the frames \texttt{[IN:CREATE\_CALL [SL:CONTACT \textrm{John} ] ]} and \texttt{[IN:CREATE\_CALL [SL:CONTACT \textrm{John Smith} ] ]} would share the same length class. The span form is also attractive in tail settings: it reduces the absolute length of the frame in open-text domains such as messaging since large utterance spans are naturally compressed.
\end{enumerate}

These forms, as we have described them, are different ``views'' of the same underlying decoupled frame representation, which allows for simple interoperability. Implementation-wise, to enable span prediction, we create \texttt{to\_span\_form()} and \texttt{from\_span\_form()} functions, which we use to encode and decode gold frames, respectively.

\subsection{Model Architecture}
\label{sec:model_arch}

We create a seq2seq, non-autoregressive semantic parser which maps an utterance $\mathbf{x} = (x_1, \cdots, x_n)$ to a frame $\mathbf{y} = (y_1, \cdots, y_m)$, where the frame is preprocessed into the span-based representation outlined above. Following \citet{nar_semantic_parsing}, we leverage the mask-predict algorithm \cite{ghazvininejad2019maskpredict}. We build our seq2seq model using an encoder, length module, and decoder; though, unlike prior approaches, we optimize the decoder to better support span prediction. Below, we elaborate on our model's components and objective:


\paragraph{Encoder.} First, we encode an utterance
\begin{equation*}
    \small
    \mathbf{h}_1, \cdots, \mathbf{h}_n = \mathrm{Encoder}(x_1, \cdots, x_n)
\end{equation*}
where the encoder is a pre-trained transformer encoder \cite{transformer}, such as RoBERTa \cite{roberta} or XLM-R \cite{conneau-2020-xlmr}.

\paragraph{Length Module.} Next, using the encoder's hidden states, a length module uses an MLP to predict the frame's length $\ell$, subsequently creating $\ell\: \times$ \texttt{[MASK]} tokens:
\begin{equation*}
\begin{split}
    \texttt{[MASK]}_1, \cdots, \texttt{[MASK]}_\ell = \\\mathrm{LengthModule}(\mathbf{h}_1, \cdots, \mathbf{h}_n)
\end{split}
\end{equation*}

\paragraph{Decoder.} Finally, using the length module's \texttt{[MASK]} tokens and encoder's hidden states, we decode a frame:
\begin{equation*}
\begin{split}
    y_1, \cdots, y_m = \mathrm{Decoder}(\texttt{[MASK]}_1, \cdots, \texttt{[MASK]}_\ell;\\ \mathbf{h}_1, \cdots, \mathbf{h}_n)
\end{split}
\end{equation*}
where the decoder is a randomly-initialized transformer decoder. Because our frames are comprised of utterance and ontology tokens, we account for this structure by equipping our decoder with a pointer-generator module \cite{see2017get, decoupled, rongali2020don}; our decoder, therefore, builds a frame by either ``copying'' utterance tokens ($y_t \in V_\textrm{cpy}$) or ``generating'' ontology tokens ($y_t \in V_\textrm{gen}$).

We critically adjust $V_\textrm{cpy}$ because our span-based frames consist of utterance \textit{index} tokens as opposed to \textit{text} tokens. Therefore, we compute the maximum index $i'$ across all utterances, and initialize $V_\textrm{cpy} = \{i : 0 \le i \le i'\}$. This setup naturally supports span prediction by enabling the decoder to unambiguously reference the endpoints of utterance spans. Furthermore, because the number of index tokens (e.g., 100) is typically far smaller than the number of text tokens (e.g., 50-200K), our parser uses substantially less memory during train- and test-time.

\paragraph{Model Objective.} Our model requires inference at two stages; the length module predicts the frame's length $\ell$ and, conditioned on this length, the decoder predicts the frame $\mathbf{y}$. As such, we jointly optimize for two objectives---a \textbf{length loss} $\mathcal{L}_\textrm{length} = \mathrm{NLL}(\ell^*, \ell) + \beta_1 \mathrm{LS}(\ell)$ and \textbf{label loss} $\mathcal{L}_\textrm{label} = \mathrm{NLL}(\mathbf{y}^*, \mathbf{y}) + \beta_2 \mathrm{LS}(\mathbf{y})$---for training the length module and decoder, respectively; though, both backpropagate to the encoder. For both objectives, we compute the negative log likelihood loss between the targets ($\ell^*$, $\mathbf{y}^*$) and predictions ($\ell$, $\mathbf{y}$) and, additionally, regularize over-confident predictions via label smoothing \cite{pereyra2017regularizing}. Empirically, we find the length loss converges quickly, so we control its influence using a scalar mixing parameter $\lambda_1$. We define the loss function as $\mathcal{L}_\textrm{NAR} = \mathcal{L}_\textrm{label} + \lambda_1 \mathcal{L}_\textrm{length}$.


We also integrate R3F \cite{RXF}, a trust-region-based algorithm which maintains the generalizability of pre-trained representations. More details are described in Appendix \S \ref{appendix:r3f}.

\paragraph{Beam Search.} During inference, the length module predicts the top $k$ lengths and we decode a parse for each length in parallel. This is similar to beam search in autoregressive decoding as we get $k$ candidates from our model during inference; however, unlike autoregressive generation, mask-predict, non-autoregressive models are only capable of generating a single beam per target length \cite{ghazvininejad2019maskpredict}.

\section{Experimental Setup}

Our goal is to benchmark span pointer networks on three axes: \textbf{quality}, \textbf{generalizability}, and \textbf{resources}. We benchmark quality by inspecting exact match (EM) on semantic parsing datasets, generalizability by probing cross-domain and cross-lingual performance, and resources by measuring latency and memory usage during inference.

\subsection{Datasets}

We experiment with the following task-oriented semantic parsing datasets: TOP (Task Oriented Parsing) \cite{gupta2018semantic}, TOPv2 \cite{chen-2020-topv2}, and MTOP (Multilingual Task Oriented Parsing) \cite{li-2020-mtop}. TOP evaluates parsers' abilities to produce nested frames across the event and navigation domains. In comparison, TOPv2 consists of both linear and nested frames and extends TOP to the alarm, messaging, music, navigation, timer, and weather domains. While TOP and TOPv2 consist entirely of utterances in English, MTOP provides gold translations in Spanish, French, German, Hindi, and Thai, making it useful for evaluating parsers in multilingual settings.

\subsection{Evaluation}

\paragraph{Quality.} When benchmarking our parser on task-oriented semantic parsing datasets, we primarily evaluate the exact match (EM) between predicted and gold frames. Note that, because our parser produces index tokens as opposed to text tokens in leaf arguments, we post-process predicted frames, mapping index $\rightarrow$ text tokens.

\paragraph{Generalizability.} Recent work evaluates the generalizability of task-oriented semantic parsers when the data distribution changes \cite{chen-2020-topv2,li-2020-mtop}. Similarly, we evaluate how generalizable our parser is in both cross-domain and cross-lingual settings, largely to understand whether span prediction is a fundamentally good paradigm for non-autoregressive parsing.

We conduct cross-domain experiments on TOPv2 \cite{chen-2020-topv2}. Our experiments proceed in two stages: we fine-tune on a high-resource, source dataset, then fine-tune on a low-resource, target dataset. Following \citet{chen-2020-topv2}, we use the alarm, event, messaging, music, navigation, and timer domains as the source dataset and reminder and weather as (separate) target datasets. We also compare our parser's EM at different levels of samples per intent slot (SPIS)\footnote{By performing random sampling with $k$ SPIS, we ensure that at least $k$ samples exist for each intent/slot label, ensuring full coverage over the ontology space. Empirically, the total number of examples sampled is 10 $\times$ $k$ SPIS, so, for example, we can roughly expect 100 unique examples with 10 SPIS. See \citet{chen-2020-topv2} for more details.}, \cite{chen-2020-topv2}, a metric used to subsample target datasets; specifically, we compute EM at 10, 25, 50, 100, and 1000 SPIS.

In addition, we conduct cross-lingual experiments on MTOP \cite{li-2020-mtop}. We primarily evaluate the zero-shot capabilities of our parser when transferred to a different language. Specifically, we create a multilingual version of our parser using XLM-R \cite{conneau-2020-xlmr} as the pre-trained encoder. Then, we train our parser on English samples and test it on non-English samples (i.e., Spanish, French, German, Hindi, or Thai).

\paragraph{Resources.} Production-ready conversational assistants typically have stringent resource requirements; if deployed in a real-world setting, how many resources would our parser require? We focus on latency (ms) and memory consumption (mb), as these are standard metrics practitioners track during deployment. To begin, we export our model with TorchScript \cite{NEURIPS2019_PyTorch} and dynamically quantize it to compress its weights. Our benchmark server uses a Intel Xeon CPU with 256GB RAM, and we restrict our models to run on 4 threads. For latency, we run our TorchScript model on a CPU against the TOPv2 dev split \cite{chen-2020-topv2}. And, for memory, we track the amount of memory allocated or released during the execution of the model's operators using the PyTorch profiler.\footnote{\url{https://pytorch.org/tutorials/recipes/recipes/profiler\_recipe.html}} Unless specified otherwise, we report the 99th percentile (P99) latency and for memory consumption we measure maximum memory consumption on the longest parse.

\subsection{Systems for Comparison}

We chiefly compare against three task-oriented semantic parsing models, which combine a seq2seq transformer with a pointer-generator-based decoder and cover both autoregressive (AR) and non-autoregressive (NAR) training:

\paragraph{BART (AR).} BART is a seq2seq transformer combining a transformer encoder and autoregressive transformer decoder, and is pre-trained with a denoising autoencoder objective on monolingual corpora \cite{lewis2019bart}. For task-oriented semantic parsing, \citet{decoupled} shows BART achieves state-of-the-art EM on multiple datasets.

\paragraph{RoBERTa AR.} RoBERTa is a transformer encoder pre-trained using an optimized BERT objective on monolingual corpora \cite{roberta}. Because transduction-based semantic parsing requires a seq2seq model, \cite{decoupled, rongali2020don} combine a RoBERTa encoder and an autoregressive, randomly-initialized transformer decoder (3L, 768/1024H, 16/24A).

\paragraph{RoBERTa NAR.} Unlike RoBERTa AR, RoBERTa NAR assumes strong independence assumptions during decoding, using the mask-predict algorithm to enable non-autoregressive generation \cite{ghazvininejad2019maskpredict}. We use the framework outlined in \citet{nar_semantic_parsing}, creating a seq2seq transformer with a RoBERTa encoder, MLP length module, and a non-autoregressive, randomly-initialized transformer decoder. Empirically, we find RoBERTa NAR relies on the transformer encoder more, so we make the transformer decoder shallower (1L, 768/1024H, 16/24A); these observations are also consistent with the recent ``deep encoder, shallow decoder'' findings in efficient machine translation \cite{Kasai2020DeepES}.

Because span pointer networks build on top of RoBERTa NAR, subsequently modifying the frame representation and model architecture to be span-based, we denote our parser as RoBERTa NAR + Span Pointer in our experiments.

\section{Results and Discussion}

\begin{table}[t]
\small
\centering
\begin{tabular}{lrr}
    \toprule
    Model & TOPv2 & TOP \\
    \midrule
    \multicolumn{3}{l}{Type: Autoregressive Models (Prior)} \\
    \midrule
    BERT$^\diamondsuit_\textrm{LARGE}$ & --- & 83.13  \\
    RoBERTa$^\diamondsuit_\textrm{LARGE}$ & --- & 86.67  \\
    RoBERTa$^\heartsuit_\textrm{LARGE}$ & --- & 84.52  \\
    BART$^\heartsuit_\textrm{LARGE}$ & --- & \textbf{87.10} \\
    \midrule
    \multicolumn{3}{l}{Type: Autoregressive Models (Ours)} \\
    \midrule
    RoBERTa$_\textrm{BASE}$ & 86.62 & 83.17 \\
    RoBERTa$_\textrm{LARGE}$ &  86.25  & 82.24  \\
    BART$_\textrm{BASE}$ & 86.73 & 84.33  \\
    BART$_\textrm{LARGE}$ & \textbf{87.48} & \textbf{85.71} \\
    \midrule
    \multicolumn{3}{l}{Type: Non-Autoregressive Models (Ours)} \\
    \midrule
    RoBERTa$_\textrm{BASE}$ & 85.78 & 82.37  \\
    \quad+ Span Pointer & 86.93 & 84.45  \\
    RoBERTa$_\textrm{LARGE}$ & 86.25 & 83.40 \\
    \quad+ Span Pointer & \textbf{87.37} & \textbf{85.07}  \\
    \bottomrule
\end{tabular}
\caption{\textbf{ EM performance on TOPv2 and TOP.} We close the quality gap between non-autoregressive and autogressive parsing; our span pointer parser matches the BART parser, despite parallel decoding. Note that, unlike prior work, we minimize the amount of hyperparameter tuning, therefore absolute EM scores are not directly comparable. $^\diamondsuit$\citet{rongali2020don}; $^\heartsuit$\citet{decoupled}}
\label{tab:results}
\end{table}

\begin{table*}[h]
\small
\centering
\setlength{\tabcolsep}{4pt}
\begin{tabular}{lrrrrrrrrrrrr}
\toprule
 & \multicolumn{6}{c}{Weather Domain (SPIS)} & \multicolumn{6}{c}{Reminder Domain (SPIS)} \\
 \cmidrule(lr){2-7} \cmidrule(lr){8-13}
 & 10 & 25 & 50 & 100 & 1000 & Avg & 10 & 25 & 50 & 100 & 1000 & Avg \\ \midrule
\multicolumn{13}{l}{Type: Autoregressive Models} \\ \midrule
RoBERTa AR & 69.71 & \textbf{74.90} & \textbf{77.02} & 78.69 & \textbf{86.36} & 77.34 & 49.38 & 56.98 & 62.18 & 69.17 & 78.48 & 63.24 \\
BART AR & \textbf{73.34} & 73.55 & 76.58 & \textbf{79.16} & 86.25 & \textbf{77.78} & \textbf{49.75} & \textbf{63.31} & \textbf{69.60} & \textbf{72.01} & \textbf{80.82} & \textbf{67.10} \\ \midrule
\multicolumn{13}{l}{Type: Non-Autoregressive Models}  \\ \midrule
RoBERTa NAR & 59.01 & 72.12 & 73.41 & \textbf{78.48} & 87.42 & 74.09 & 33.90 & 40.19 & 49.87 & 54.76 & 76.71 & 51.09 \\
\quad+ Span Pointer & \textbf{72.03} & \textbf{74.74} & \textbf{74.85} & 78.14 & \textbf{88.47} & \textbf{77.65} & \textbf{48.27} & \textbf{60.55} & \textbf{68.11} & \textbf{73.19} & \textbf{80.44} & \textbf{66.11} \\
\bottomrule
\end{tabular}
\caption{\textbf{Cross-domain generalizability experiments on TOPv2, comparing EM with SPIS.} We perform high-resource fine-tuning on multiple source domains, then low-resource fine-tuning on two target domains, weather and reminder, separately. Each target domain consists of multiple subsets for fine-tuning; these are created by randomly sampling $k$ samples per intent slot (SPIS) \cite{chen-2020-topv2}.}
\label{tab:sample-efficiency}
\end{table*}

\subsection{Quality}

Table \ref{tab:results} shows EM results on TOPv2 and TOP. We see that, with our span pointer formulation, we improve upon prior non-autoregressive parsers by +1-2 EM. These gains are consistent across pre-trained encoders, particularly RoBERTa$_\textrm{BASE}$, indicating real-world applicability where smaller model architectures are preferred. We also compare against strong autoregressive parsers. Though non-autoregressive parsers typically underperform autoregressive parsers, often requiring additional tricks like distillation \cite{gu2017non, ghazvininejad2019maskpredict, zhou2019understanding}, we see that span pointer networks largely close this quality gap. The non-autoregressive RoBERTa$_\textrm{LARGE}$ span pointer parser is comparable to the autoregressive BART$_\textrm{LARGE}$ parser, the current state-of-the-art parser, despite being non-autoregressive.

\subsection{Generalizability}

We investigate our parser's generalizability: how does it perform when the data distribution suddenly changes? We experiment with two types of distribution shifts, setting up cross-domain and cross-lingual experiments.

For cross-domain experiments, we conduct high-resource fine-tuning on multiple source domains, then low-resource fine-tuning on two target domains: weather and reminder. Table \ref{tab:sample-efficiency} lists individual EM scores at 10, 25, 50, 100, and 1000 SPIS as well as an average EM score across all SPIS values. Although there is a large EM gap between RoBERTa AR and RoBERTa NAR, our span-based, non-autoregressive parser closes this gap, matching or exceeding the EM both the RoBERTa and BART autoregressive parsers in most cases. Our parser notably does well on reminder, which is a more challenging domain given its large ontology and high compositionality \cite{chen-2020-topv2}. 

For cross-lingual experiments, we perform zero-shot evaluations where a multilingual version of our parser (using XLM-R as the pre-trained encoder) is trained on English samples and tested on non-English samples. Table \ref{tab:xl} shows EM scores for five such zero-shot settings: English $\rightarrow$ Spanish, French, German, Hindi, and Thai. Our span-based, non-autoregressive parser consistently outperforms both non-autoregressive and autoregressive baselines, achieving +14-15 average EM. We attribute the improvement in our model to be due to the length predictor indirectly predicting frame syntax which is language-agnostic. For example, \texttt{[IN:GET\_WEATHER [SL:LOCATION \textrm{entertainment center} ] ]} and \texttt{[IN:GET\_WEATHER [SL:LOCATION \textrm{centro de entretenimiento} ] ]} are English-Spanish parallel samples; using the canonical form, the length discrepancy makes it challenging for mask-predict models to learn the association, but by using the span form, our model is able to seamlessly bridge this gap. Our findings are consistent with prior work showing that non-autoregressive modeling can be beneficial in zero-shot multilingual settings \cite{Zhu2020DontPI}.


\begin{table*}[]
\small
\centering
\begin{tabular}{lrrrrrrr}
\toprule
 & \multicolumn{7}{c}{Zero-Shot Evaluation} \\
\cmidrule(lr){3-8}
 & \texttt{en} & \texttt{en$\rightarrow$es} & \texttt{en$\rightarrow$fr} & \texttt{en$\rightarrow$de} & \texttt{en$\rightarrow$hi} & \texttt{en$\rightarrow$th} & Avg \\
\midrule
XLM-R$_\textrm{BASE}$ NAR & 78.3 & 35.2 & 32.2 & 23.6 & 18.1 & 16.7 & 25.2 \\
\quad + Span Pointer & 83.0 & 51.2 & 51.4 & 42.0 & 29.6 & 27.3 & 40.3 \\
XLM-R$_\textrm{LARGE}$ NAR & 80.5 & 50.9 & 51.5 & 38.7 & 31.6 & 22.8 & 39.1 \\
\quad + Span Pointer & \textbf{84.5} & \textbf{60.4} & \textbf{63.1} & \textbf{56.2} & \textbf{41.2} & \textbf{41.7} & \textbf{52.5} \\
XLM-R$_\textrm{LARGE}$ AR$^\diamondsuit$ & 83.9 & 50.3 & 43.9 & 42.3 & 30.9 & 26.7 & 38.8 \\
\bottomrule
\end{tabular}
\caption{ \textbf{Cross-lingual generalizability experiments on MTOP, comparing EM across zero-shot setups.} We perform zero-shot experiments where we fine-tune a parser on English (\texttt{en}), then evaluate it a non-English language---Spanish (\texttt{es}), French (\texttt{fr}), German (\texttt{de}), Hindi (\texttt{hi}), and Thai (\texttt{th})---without fine-tuning. Average EM (Avg) is taken over the five non-English languages. $^\diamondsuit$\citet{li-2020-mtop}}
\label{tab:xl}
\end{table*}

\subsection{Resources}
\begin{table*}[h]
\small
\centering
\begin{tabular}{@{}lrrrrrr@{}}
\toprule
\multirow{2}{*}{Model} & \multirow{2}{*}{EM (NQ / Q) $\uparrow$} &  \multirow{2}{*}{\# Params $\downarrow$} & \multicolumn{2}{c}{Latency (ms) $\downarrow$} &  \multicolumn{2}{c}{Memory (mb) $\downarrow$} \\ \cmidrule(lr){4-5} \cmidrule(lr){6-7}
& & &  $k$ = 1 & $k$ = 5 & $k$ = 1 & $k$ = 5 \\ \midrule
BART$_\textrm{BASE}$ & \textbf{87.00} / 84.67 & 221M &  1,143 & 2,131  & 93 & 437\\
RoBERTa$_\textrm{BASE}$ AR & 86.51  / 86.26 & 183M & 1,154 & 3,983 & 582 &  2,390 \\
RoBERTa$_\textrm{BASE}$ NAR & 85.78 / 85.60 & 142M & 149 & 680 & 52 & 211  \\
\quad+ Span Pointer (Index Form) & 86.12 / 86.05 & \textbf{134M} & 145 & \textbf{199} & 23 & \textbf{35} \\
\quad+ Span Pointer (Span Form) & \textbf{86.99} / \textbf{86.80} & \textbf{134M} & \textbf{134} & 208 & \textbf{22} & 42 \\
\bottomrule
\end{tabular}
\caption{\textbf{ Latency and memory benchmarking on TOPv2.} We report EM (non-quantized / quantized) in addition to, number of parameters, latency (ms) and memory (mb) at variable beam sizes ($k \in \{\textrm{1}, \textrm{5}\}$). Our span pointer parser achieves the best quantized EM and resource reductions over both autoregressive and non-autoregressive baselines.}
\label{tab:resources}
\end{table*}

Our results above indicate that the RoBERTa$_\textrm{BASE}$ NAR + Span Pointer parser is qualitatively strong, despite its smaller transformer encoder, but we have not yet investigated its resource requirements. 

Using the base variants of BART AR, RoBERTa AR, and RoBERTa NAR with variable beam sizes ($k$ = 1 and 5), we quantize each model to compress its weights, then benchmark latency and memory on a CPU-based server.\footnote{We use 1 decoder layer for RoBERTa AR and RoBERTa NAR to ensure fair comparison, though note RoBERTa AR typically requires 3 decoder layers for best quality.} Table \ref{tab:resources} shows these results; from here, we make a couple of observations. First, our quantized parser achieves both higher EM and lower latency/memory compared to autoregressive BART and RoBERTa parsers. When compared to BART, our parser cuts latency by 8.5$\times$ and 10$\times$ at $k$ = 1 and 5, respectively, which is largely due to high parallelism during generation. Second, we also improve upon RoBERTa NAR, the base non-autoregressive parser, which we attribute to the variants of the decoupled frame representation we explore. Specifically, index form targets memory by avoiding the need to store large subword vocabularies and span form targets latency by minimizing the size of leaf arguments in frames. Overall, with $k$ = 5, our parser cuts latency by 3.2$\times$ and memory by 4.9$\times$, indicating its usability in real-world settings.

\section{Analysis}

Having evaluated span pointer networks on several axes, we now turn towards understanding the driving factors behind its performance.

\subsection{Model Ablations}
\begin{table}[t]
\small
\centering
\begin{tabular}{lrr}
\toprule
Model (RoBERTa$_\textrm{BASE}$) & TOPv2 & TOP \\
\midrule
Span Pointer & \textbf{86.99} & \textbf{84.74} \\
\midrule
\multicolumn{3}{l}{Ablation: Representation + Architecture} \\
\midrule
\quad- Span Form & 85.82 & 82.99 \\
\quad\quad- Index Form$^*$ & 85.45 & 82.50 \\
\midrule
\multicolumn{3}{l}{Ablation: Fine-tuning} \\
\midrule
\quad- R3F & 86.73 & 83.64 \\
\bottomrule
\end{tabular}
\caption{ Model ablation experiments, comparing EM scores when the representation + architecture and fine-tuning components are modified. $^*$To isolate the index form's contribution, we use the canonical frame representation (\S \ref{sec:index_span}) and standard non-autoregressive architecture (\S \ref{sec:model_arch}).}
\label{tab:ablation}
\end{table}

Our parser critically relies on three components: frame representation, model architecture, and fine-tuning. In Table \ref{tab:ablation}, we present a series of ablation experiments to isolate the contribution of each component. First, we consider our span pointer network without the span and index forms (\S \ref{sec:index_span}). Here, to remove index form, specifically, we also have to undo our architectural changes (\S \ref{sec:model_arch}), making the resulting model identical to RoBERTa NAR. Our complete parser achieves higher EM, suggesting our representation and architecture changes, together, have a positive impact on performance. Second, we consider our span pointer network with and without R3F fine-tuning. While our parser achieves better EM with R3F, especially on TOP, these results indicate our method is not entirely contingent on better fine-tuning.

\subsection{Span Prediction and Target Length}

\begin{table}[h]
\small
\centering
\begin{tabular}{lrr}
\toprule
 & Canonical & Span \\ \midrule
\# Length Classes &  47 & 20 \\
Mean Frame Length per Frame &  1.94 & 1.0 \\
Mean Frame Length &  10.50 & 9.71 \\
Max Frame Length &  62 & 58 \\ 
\bottomrule
\end{tabular}
\caption{Length distribution statistics of TOPv2 \cite{chen-2020-topv2} frames when comparing canonical and span forms.}
\label{table:topv2_length}
\end{table}

In Table \ref{table:topv2_length} we show the length characteristics of both the canonical and span form on TOPv2 \cite{chen-2020-topv2}. Span form leads to a tighter and more consistent distribution as it only takes 1 length class to represent each unique frame, leading to many fewer length classes as well.

\subsection{Frame Syntax vs. Semantics}

\begin{table}[h]
\centering
\small
\begin{tabular}{lrr}
\toprule
 & \multicolumn{2}{c}{Decoupled Frame} \\
 \cmidrule(lr){2-3}
 & Syn Only & Syn + Sem \\
\midrule
\multicolumn{3}{l}{Type: Base Models} \\
\midrule
BART & 88.29 & \textbf{86.10} \\
RoBERTa AR & 88.11 & 86.29 \\
RoBERTa NAR & \textbf{88.53} & 84.63 \\
\midrule
\multicolumn{3}{l}{Type: Large Models} \\
\midrule
BART & 89.01 & \textbf{87.47} \\
RoBERTa AR & 88.40 & 85.94 \\
RoBERTa NAR & \textbf{89.32} & 85.66 \\
\bottomrule
\end{tabular}
\caption{ EM performance in ``syntax only'' (Syn Only) and ``syntax + semantics'' (Syn + Sem) settings. Here, ``syntax only'' and ``syntax + semantics'' refer to decoupled frames without and with leaf arguments, respectively. In the ``syntax only'' setting, the non-autoregressive parser outperforms, suggesting the generation of leaf arguments is a major bottleneck.}
\label{tab:syntax-and-semantics}
\end{table}

Our initial motivation for modifying non-autoregressive, mask-predict parsers stems from the argument that, when shifting to span prediction from text generation, a length module only needs to predict frame syntax as opposed to both frame syntax and semantics. We make the implicit assumption that frame syntax is \textit{easier} to learn than frame semantics. Though this assumption is supported by recent work fine-tuning span-based transformers for task-oriented semantic parsing \cite{desai-2021-dianosing}, we now devise an experiment to explicitly test this hypothesis.

We refer to a regular decoupled frame \texttt{[IN:SEND\_MESSAGE [SL:CONTENT [} I'll be there at 6pm \texttt{] ] ]} as ``syntax + semantics'', while a decoupled frame without leaf arguments \texttt{[IN:SEND\_MESSAGE [SL:CONTENT ] ]} as ``syntax only''. Here, the ``syntax only'' frame mimics a constituency tree, representing coarse-grained \textit{structure} rather than fine-grained \textit{meaning}. Using TOPv2, we create two training sets, one with ``syntax only'' frames and one with ``syntax + semantics'' frames. Table \ref{tab:syntax-and-semantics} shows EM scores when fine-tuning BART, RoBERTa AR, and RoBERTa NAR parsers on these training sets.

In the ``syntax only'' setting, RoBERTa NAR slightly outperforms both BART and RoBERTa AR, while in the ``syntax + semantics'' setting, it consistently lags behind. Recall that, operationally, the main difference between these settings is that, in ``syntax only'', RoBERTa NAR's length module is only responsible for frame structure, while in ``syntax + semantics'', it is responsible for both frame structure and arguments. Our results suggest that the main difficulty in non-autoregressive modeling is handling leaf arguments, so simplifications of frame semantics (i.e., our proposed span form) are likely to improve quality. 

\section{Conclusion}

In this work, we present span pointer networks for task-oriented semantic parsing: seq2seq, non-autoregressive models which focus on span prediction as opposed to text generation. Our approach requires creating a span-based frame representation and model architecture, which, together, enable simple and consistent non-autoregressive modeling. We benchmark our parser on three axes---quality, generalizability, and resources---and demonstrate real-world applicability by improving upon both non-autoregressive and autoregressive baselines. Future work can extend our approach to work with discontinuous spans, as is important in session-based \cite{decoupled} and free word order \cite{li-2020-mtop} modeling.

\bibliography{anthology,custom}
\bibliographystyle{acl_natbib}

\appendix
\newpage

\section{Training Details}
\label{sec:training-details}

\paragraph{Implementation.} For RoBERTa AR and RoBERTa NAR, we reference the open-source implementation of \citet{nar_semantic_parsing} in PyText \cite{aly2018pytext}. For BART AR, we reference its implementation in fairseq \cite{ott2019fairseq}. For experimentation on TOPv2 \cite{chen-2020-topv2}, we use 8 16GB GPUs, due to its large dataset size, and for TOP \cite{gupta2018semantic}, we use 1 16GB GPU. For BART$_\textrm{LARGE}$, we use 1 32GB GPU.

\paragraph{Hyperparameters.} We determine model hyperparmeters by sweeping across the pre-defined ranges in Appendix \ref{appendix:hp}. For optimization of RoBERTa and XLM-R, specifically, we use Adam \cite{kingma2014adam} and learning rate schedulers; autoregressive models use exponential learning decay and non-autoregressive models use decay on plateau, following the guidance in \citet{nar_semantic_parsing}. For optimization of BART, we use stochastic weight averaging (SWA) \cite{swa} and LAMB \cite{lamb} following \citet{decoupled}.

Though fine-tuning with R3F typically yields the best results on generation tasks \cite{RXF}, we treat it as a hyperparameter, and report the max performance both with and without R3F. We also perform an ablation in Appendix \ref{sec:r3f_ablation} to demonstrate its importance.

\section{Better Fine-tuning with R3F}
\label{appendix:r3f}

We also integrate R3F \cite{RXF} into our proposed Span Pointer Networks, a trust-region-based algorithm which maintains the generalizability of pre-trained representations. Following \citet{RXF}, we create auxiliary terms for both objectives

\begin{equation*}
\begin{split}
    \mathcal{L}_\textrm{R3F-length} = \mathrm{KL}_\mathrm{S}(\mathrm{LengthModule}(x) || \\ \mathrm{LengthModule}(x + z) \\
    \mathcal{L}_\textrm{R3F-label} = \mathrm{KL}_\mathrm{S}(\mathrm{Decoder}(x) || \\ \mathrm{Decoder}(x + z) \\
\end{split}
\end{equation*}

where $\mathrm{KL}_\mathrm{S}$ represents the symmetric Kullback-Leibler divergence between a regular and noised input and $z \sim \mathcal{U}(-\sigma, \sigma)$. Then, we add these terms to the length and label losses, respectively, resulting in a smoother objective:

\begin{equation*}
    \mathcal{L}_\textrm{R3F-NAR} = \mathcal{L}_\textrm{NAR} + \lambda_2 \mathcal{L}_\textrm{R3F-length} \\+ \lambda_3 \mathcal{L}_\textrm{R3F-label}
\end{equation*}

\subsection{R3F Ablations}
\label{sec:r3f_ablation}

In Table \ref{table:r3f_ablation} we show the impact of leveraging R3F \cite{RXF} across our baseline model architectures. We see consistent improvements across generation strategies (autoregressive and non-autoregressive) for our BART and RoBERTa-based parsers. 

\begin{table}[h]
\small
\centering
\begin{tabular}{llll}
\toprule
Model & TOPv2 & TOP \\
\midrule
\multicolumn{3}{l}{Type: Autoregressive Models} \\
\midrule
RoBERTa$_\textrm{BASE}$ & 86.29 & 82.08 \\
\quad + R3F & 86.62 & 83.17 \\
RoBERTa$_\textrm{LARGE}$ & 85.94 & 82.64  \\
\quad + R3F &  86.25  & 82.24 \\
BART$_\textrm{BASE}$ & 86.10 & 83.62  \\
\quad + R3F & 86.73 & 84.33 \\
BART$_\textrm{LARGE}$ & \textbf{87.48} & 85.53 \\
\quad + R3F & 86.96 & \textbf{85.71} \\
\midrule
\multicolumn{4}{l}{Type: Non-Autoregressive Models} \\
\midrule
RoBERTa$_\textrm{BASE}$ & 85.13 & 82.06  \\
\quad + R3F & 85.78 & 82.37 \\
RoBERTa$_\textrm{LARGE}$ & 85.93 & 82.57   \\
\quad + R3F  &\textbf{86.25} & \textbf{83.40} \\
\bottomrule
\end{tabular}
\caption{EM scores of autoregressive and non-autoregressive parsers when leveraging R3F-based fine-tuning \cite{RXF}.}
\label{table:r3f_ablation}
\end{table}

\section{Autoregressive Span Pointer Networks}

\begin{table}[h]
\small
\centering
\begin{tabular}{llll}
\toprule
Model & TOPv2 & TOP \\
\midrule
\multicolumn{3}{l}{Type: Autoregressive Models} \\
\midrule
RoBERTa$_\textrm{BASE}$ & 86.62 & 83.17 \\
\quad + Span Pointer & 86.72 & 84.54  \\
RoBERTa$_\textrm{LARGE}$ & 86.25  & 82.24  \\
\quad + Span Pointer & 86.86  & 84.66  \\
\midrule
\multicolumn{3}{l}{Type: Non-Autoregressive Models} \\
\midrule
RoBERTa$_\textrm{BASE}$ & 85.78 & 82.37  \\
\quad+ Span Pointer & 86.93 & 84.45  \\
RoBERTa$_\textrm{LARGE}$ & 86.25 & 83.40 \\
\quad+ Span Pointer & \textbf{87.37} & \textbf{85.07}  \\
\bottomrule
\end{tabular}
\caption{EM scores for auto-regressive Span Pointer Networks}
\label{table:ar_span_pointer}
\end{table}

In table \ref{table:ar_span_pointer} we present the results of auto-regressive span pointer networks. We see that our formulation of leveraging span prediction helps in the autoregressive setting as well as the non-autoregressive setting. However, we see larger improvements in non-autoregressive span pointer networks, even surpassing the auto-regressive variants despite being non-autoregressive.

\section{Hyperparameters}
\label{appendix:hp}

We specify the details of our hyperparameter sweeps in two tables below. Table \ref{table:hp_ar} specifies the auto-regressive hyperparameters for BART and RoBERTa models. Table \ref{table:hp_nar} we specify the non-autoregressive hyperparameters for RoBERTa and XLM-R models used in the paper. For the auto-regressive and non-autoregressive models, the models were hand tuned for initial parameters and the final numbers reported were based on a hyperparameter sweep with a total of 36 runs. For BART \cite{lewis2019bart} based models, we keep optimization parameter consistent with \citet{decoupled} rather than tuning the model ourselves. For our non-autoregressive models, both the baseline \cite{nar_semantic_parsing} and our proposed span pointer parser, we leverage the exact same hyperparameter sweep to ensure a fair comparison.

\begin{table}[h]
\small
\centering
\begin{tabular}{@{}lrrrr@{}}
\toprule
Parameter & Value \\
\midrule
R3F $\lambda$ & Swept over\\
Noise & Uniform \\
$\epsilon$ & 0.00001 \\
\bottomrule
\end{tabular}
\caption{R3F constant hyperparameters.}
\label{table:hp_r3f}
\end{table}

\begin{table}[h]
\small
\centering
\begin{tabular}{@{}lrrrr@{}}
\toprule
Parameter & Value \\
\midrule
SWA LR & 0.0002 \\
Start Step & 18K \\
Frequency & 230 \\
\bottomrule
\end{tabular}
\caption{Stochastic Weight Averaging (SWA) hyperparameters used for BART fine-tuning.}
\label{table:hp_swa}
\end{table}

\begin{table*}[h]
\small
\centering
\begin{tabular}{@{}lrrrr@{}}
\toprule
\multirow{ 2}{*}{Parameter} & \multicolumn{4}{c}{Autoregressive Models} \\
\cmidrule(lr){2-5}
 & BART$_\textrm{BASE}$ & BART$_\textrm{LARGE}$  & RoBERTa$_\textrm{BASE}$  & RoBERTa$_\textrm{LARGE}$    \\ \midrule
Optimizer & SWA-Lamb & SWA-Lamb & Adam & Adam \\
Learning Rate Scheduler & Exp-LR($\gamma$=0.95) & Exp-LR($\gamma$=0.95) & Exp-LR($\gamma$=[0.5, 0.99]) & Exp-LR($\gamma$=[0.5, 0.99]) \\
Learning Rate     &   [5e-7, 5e-3] & [5e-7, 5e-3] &  [0.000001, 0.001]    & [0.000001, 0.001]  \\
Batch Size  & 16  &  16    & \{4,8,16\}    & \{4,8\} \\
R3F $\lambda$ & 0.01 &   0.01    &  0.01    & 0.01 \\
\# GPU & 1 & 1 & \{1,8\} & \{1,8\} \\
GPU Memory 32GB & 16GB  & 16GB  & 16GB  & 16GB \\
\bottomrule
\end{tabular}
\caption{Hyperparameter values for autoregressive model architectures.}
\label{table:hp_ar}
\end{table*}

\begin{table*}[h]
\small
\centering
\begin{tabular}{@{}lrrrrrr@{}}
\toprule
\multirow{2}{*}{Parameter} & \multicolumn{3}{c}{Non-Autoregressive Models}& \\
\cmidrule(lr){2-5}
 & RoBERTa$_\textrm{BASE}$ & RoBERTa$_\textrm{LARGE}$ & XLM-R$_\textrm{LARGE}$  \\ \midrule
Optimizer &  \multicolumn{3}{c}{Adam}  \\
Learning Rate Scheduler &  \multicolumn{3}{c}{ReduceLR} \\
Learning Rate & \multicolumn{3}{c}{[0.0002, 0.000002]} \\
Batch Size & \{32, 64\} & \{8, 16\} & 16  \\
Length Loss $\lambda_1$ & \multicolumn{3}{c}{[0.1, 1.0]}  \\
R3F $\lambda_\textrm{label}$ ($\lambda_2$) & \multicolumn{3}{c}{0.001}  \\
R3F $\lambda_\textrm{length}$ ($\lambda_3$) & \multicolumn{3}{c}{0.01} \\
$\beta_1 (\mathcal{L}_\textrm{labels})$ & \multicolumn{3}{c}{[0, 0.2]} \\
$\beta_2 (\mathcal{L}_\textrm{length})$  & \multicolumn{3}{c}{[0, 0.5]} \\
\# GPU & \multicolumn{3}{c}{\{1,8\}} \\
GPU Memory & 16GB  & 16GB  & 32GB \\

\bottomrule
\end{tabular}
\caption{Hyperparameter values for non-autoregressive model architectures.}
\label{table:hp_nar}
\end{table*}

\end{document}